\documentclass{article}

\usepackage{microtype}
\usepackage{subcaption}     % for subfigures
\usepackage{graphicx}
\usepackage{booktabs} % for professional tables
\usepackage{dsfont}      % for Reals    

\usepackage{hyperref}

\usepackage[accepted]{icml2024}

\usepackage{amsmath}
\usepackage{amssymb}
\usepackage{mathtools}
\usepackage{amsthm}

\usepackage{import}

\usepackage[capitalize,noabbrev]{cleveref}
\usepackage{pgf, tikz}
\usepackage{pgfplots}
\usepackage{pgfplotstable}
\usepackage{pgfplots}

\usetikzlibrary{arrows, automata}
\pgfplotsset{compat=1.15}
\pgfplotsset{every axis/.append style={font=\Large}}
\pgfplotsset{every tick label/.append style={font=\Large}}
\pgfplotsset{compat=1.16}
\usepgfplotslibrary{colorbrewer}
\usepgfplotslibrary{groupplots}
\usepgfplotslibrary{colormaps}

\pgfplotsset{
  compat                 = 1.16,  % Okay for Overleaf, I guess...
  filter discard warning = false, % Suppress warnings for filtered plots
  discard if not/.style 2 args={
    x filter/.append code={
      \edef\tempa{\thisrow{#1}}
      \edef\tempb{#2}
      \ifx\tempa%
        \tempb%
      \else%
        
      \fi
    }
  },
}

\pgfplotsset{%
  every non boxed x axis/.append style={x axis line style=-},
  every non boxed y axis/.append style={y axis line style=-}
}

\theoremstyle{plain}

\theoremstyle{definition}

\theoremstyle{remark}

\def\Vcal{{\mathcal V}}
\def\Ecal{{\mathcal E}}
\def\Ncal{{\mathcal N}}

\newcommand{\graph}     {\ensuremath{G}}
\newcommand{\X}         {\ensuremath{\mathbf{X}}}
\newcommand{\E}         {\ensuremath{\mathbf{E}}}
\newcommand{\Q}         {\ensuremath{\mathbf{Q}}}
\newcommand{\K}         {\ensuremath{\mathbf{K}}}
\newcommand{\V}         {\ensuremath{\mathbf{V}}}
\newcommand{\A}         {\ensuremath{\mathbf{A}}}
\newcommand{\softmax}   {\ensuremath{\mathrm{softmax}}}
\newcommand{\norm}   {\ensuremath{\mathrm{norm}}}
\newcommand{\U}         {\ensuremath{\mathbf{U}}}
\newcommand{\Concat}   {\ensuremath{\mathrm{Concat}}}
\newcommand{\W}         {\ensuremath{\mathbf{W}}}
\newcommand{\T}         {\ensuremath{\mathbf{T}}}

\newcommand{\TLayer}   {\ensuremath{\mathrm{TLayer}}}
\newcommand{\MPNN}   {\ensuremath{\mathrm{MPNN}}}
\newcommand{\GEANet}   {\ensuremath{\mathrm{GEANet}}}
\newcommand{\FFN}   {\ensuremath{\mathrm{FFN}}}
\newcommand{\real}{\ensuremath{\mathds{R}}}

\definecolor{dark2green}{rgb}{0.1, 0.65, 0.3}
\definecolor{dark2orange}{rgb}{0.9, 0.4, 0.}
\definecolor{dark2purple}{rgb}{0.4, 0.4, 0.8}
\newcommand{\first}[1]{\textbf{\textcolor{dark2green}{#1}}}
\newcommand{\second}[1]{\textbf{\textcolor{dark2orange}{#1}}}
\newcommand{\third}[1]{\textbf{\textcolor{dark2purple}{#1}}}
\definecolor{line1}{RGB}{246,194,66}
\definecolor{line2}{RGB}{165,165,165}
\definecolor{line3}{RGB}{222,131,68}
\definecolor{line4}{RGB}{222,138,168}
\definecolor{line5}{RGB}{176,36,24}
\definecolor{line6}{RGB}{79,113,190}
\definecolor{line7}{RGB}{126,171,85}
\definecolor{bar1}{RGB}{144,193,233}
\definecolor{bar2}{RGB}{232,146,142}

\usepackage[textsize=tiny]{todonotes}

\icmltitlerunning{Graph External Attention Enhanced Transformer}

\begin{document}

\twocolumn[
\icmltitle{Graph External Attention Enhanced Transformer}



\icmlsetsymbol{equal}{*}

\begin{icmlauthorlist}
\icmlauthor{Jianqing Liang}{yyy}
\icmlauthor{Min Chen}{yyy}
\icmlauthor{Jiye Liang}{yyy}
\end{icmlauthorlist}

\icmlaffiliation{yyy}{Key Laboratory of Computational Intelligence and Chinese Information Processing of Ministry of Education, 
School of Computer and Information Technology, Shanxi University,
Taiyuan 030006, Shanxi, China}

\icmlcorrespondingauthor{Jiye Liang}{ljy@sxu.edu.cn}

\icmlkeywords{Graph External Attention, Transformer}

\vskip 0.3in
]



\printAffiliationsAndNotice{}  % leave blank if no need to mention equal contribution

\begin{abstract}
The Transformer architecture has recently gained considerable attention in the field of graph representation learning, 
as it naturally overcomes several limitations of Graph Neural Networks (GNNs) with customized attention mechanisms or positional and structural encodings. 
Despite making some progress, existing works tend to overlook external information of graphs, specifically the correlation between graphs.
Intuitively, graphs with similar structures should have similar representations.
Therefore, we propose Graph External Attention (GEA) --- a novel attention mechanism that leverages multiple external node/edge key-value units to capture inter-graph correlations implicitly.
On this basis, we design an effective architecture called Graph External Attention Enhanced Transformer (GEAET), which integrates local structure and global interaction information for more comprehensive graph representations.
Extensive experiments on benchmark datasets demonstrate that GEAET achieves state-of-the-art empirical performance.
The source code is available for reproducibility at: \url{https://github.com/icm1018/GEAET}.
\end{abstract}

\section{Introduction}
Graph representation learning has attracted widespread attention in the past few years. 
It plays a crucial role in various applications, 
such as social network analysis~\citep{social_networks2}, drug discovery~\citep{drug_discovery}, protein design~\citep{protein_design}, medical diagnosis~\citep{medical_diagnosisxxx} and so on.

Early research in graph representation learning primarily focuses on Graph Neural Networks (GNNs). 
A milestone example is GCN~\citep{GCN2016,GCN2017semi}. It performs convolution operations on the graph. 
Based on the framework of message-passing GNNs~\citep{message_passing}, GraphSage~\citep{GraphSage}, GatedGCN~\citep{GatedGCN} and GIN~\citep{GIN} adapt to complex graph data by employing different message-passing strategies.
While message-passing GNNs have recently emerged as prominent methods for graph representation learning, 
there still exist some critical limitations,  
including the limited expressiveness~\citep{GIN,morris2019weisfeiler}, over-smoothing~\citep{over_smoothing1, over_smoothing3, over_smoothing4}, over-squashing~\citep{TreeNeighbourMatch} and poor long-range dependencies.

Instead of aggregating local neighborhood, Graph Transformers (GTs) capture interaction information between any pair of nodes through a single self-attention layer. 
Some of the existing works focus on customizing specific attention mechanisms or positional encodings~\citep{GT,Graphormer,SAN,EGT,Grit}, while others combine message-passing GNNs to design hybrid architectures~\Citep{GraphTrans,SAT,GPS}. 
These methods enable nodes to interact with all other nodes within a graph, facilitating the direct modeling of long-range relations. 
This may address typical issues such as over-smoothing in GNNs.

\begin{figure}[ht]
    \newcommand{\wth}{.45}
    \centering
    {
    \hspace{8pt} \includegraphics[width=\wth\textwidth]{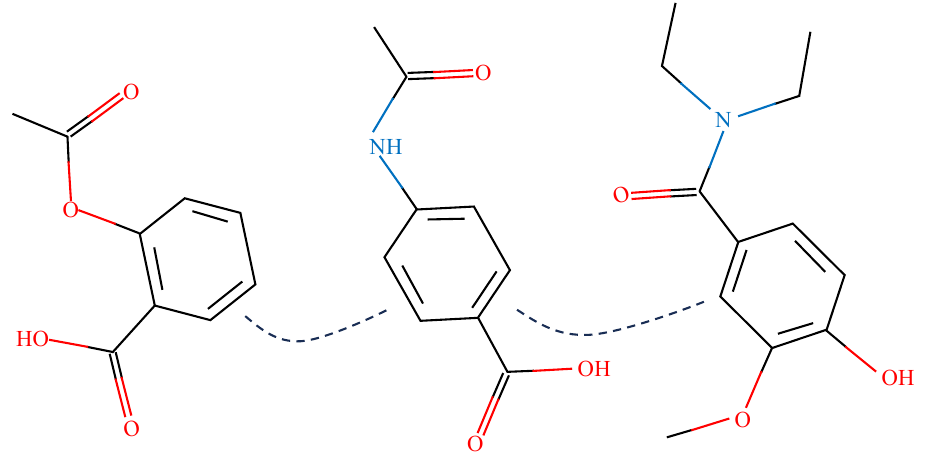}
    
    \vspace{10pt}
    }
    \caption{Three molecular graphs from the ZINC dataset are correlated to the benzene ring structure.}
    \label{fig:zinc33}
\end{figure}

While the above-mentioned methods have achieved impressive results, they are confined to internal information within the graph, neglecting potential correlations with other graphs.
In fact, strong correlations between different graphs generally exist in numerous practical scenarios, such as molecular graph data. 
Figure~\ref{fig:zinc33} shows 3 molecular graphs with a benzene ring structure. Intuitively, exploiting inter-graph correlations can improve the effectiveness of graph representation learning. 

In this work, we address the critical question of how to incorporate external information into graph representation learning. 
Our principal contribution is to introduce a novel Graph External Attention (GEA) mechanism, which implicitly learns inter-graph correlations with the external key-value units. 
Moreover, we design Graph External Attention Enhanced Transformer (GEAET), combining inter-graph correlations with both local structure and global interaction information. 
This enables the acquisition of more comprehensive graph representations, in contrast to most existing methods. 
Our contributions are listed as follows.
\begin{itemize}
    \item We introduce GEA to implicitly learn correlations between all graphs. The complexity scales linearly with the number of nodes and edges.
    \item We propose GEAET, which uses GEA to learn external information and integrates local structure and global interaction information, resulting in more comprehensive graph representations.
    \item We demonstrate that GEAET achieves state-of-the-art performance. Furthermore, we highlight the significance of GEA, emphasizing its superior interpretability and reduced dependence on positional encoding compared to self-attention.
\end{itemize}
\begin{figure*}[ht]
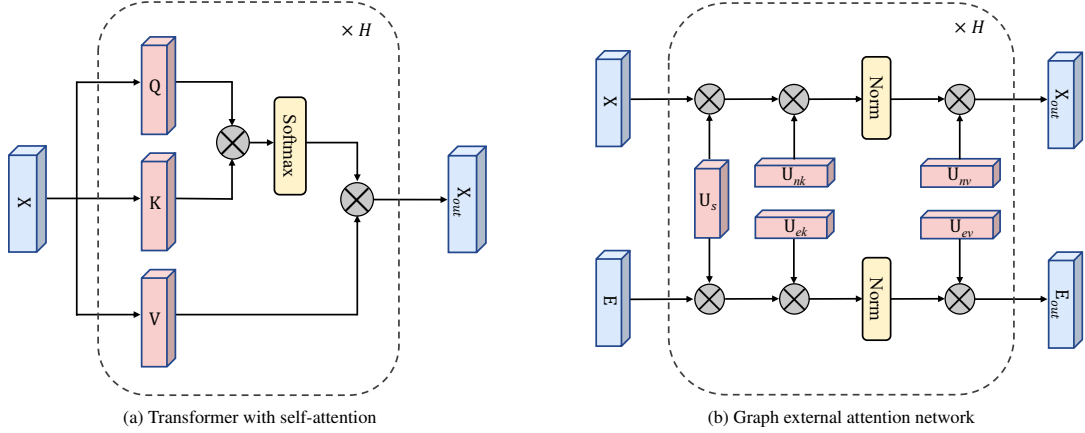

    \centering
    \resizebox{.85\textwidth}{!}{
    \begin{subfigure}[b]{0.5\textwidth}
        \centering
        \input{Figures/self}
    \end{subfigure}
    \hfill
    \hspace{1.5cm}
    \begin{subfigure}[b]{0.5\textwidth}
        \centering
        \input{Figures/external}
    \end{subfigure}
    }
       \caption{Transformer versus graph external attention network. For simplicity, we omit skip connections and FFNs.}
       \label{fig:ablation}
\end{figure*}

\section{Related Work}
\paragraph{Message-Passing Graph Neural Networks.} 
Early developments include GCN~\citep{GCN2016,GCN2017semi}, GraphSage~\citep{GraphSage}, GIN~\citep{GIN}, GAT~\citep{GAT}, GatedGCN~\citep{GatedGCN} and others. 
These methods are based on a message-passing architecture~\citep{message_passing} that generally faces the challenges of limited expressivity. 
Recent advancements include various works attempting to enhance GNNs to improve expressivity. 
Examples include some works that add features to distinguish nodes~\citep{murphy2019relational,sato2021random,qiu2018network,bouritsas2020improving,lspe}. 
Others focus on altering the message-passing rule~\citep{DGN} or modifying the underlying graph structure for message-passing~\citep{morris2019weisfeiler,bodnar2021weisfeiler} to further exploit the graph structure.

Despite achieving state-of-the-art performance, GNNs face over-smoothing and over-squashing due to their constrained receptive field. 
Over-smoothing happens when all node representations converge to a constant after deep layers, whereas over-squashing occurs when messages from distant nodes fail to propagate effectively. 
It is crucial to design new architectures beyond neighborhood aggregation to address these issues.
\paragraph{Graph Transformers.} 
The Transformer with self-attention, a dominant approach in natural language processing~\citep{Transformer,bert}, has shown competitiveness in computer vision~\citep{transformer_image}. 
Given the remarkable achievements of Transformers and their capability to address crucial challenges of GNNs, GTs have been proposed, attracting increasing attention.
Existing works primarily focus on designing tailored attention mechanisms or positional and structural encodings, or combining message-passing GNNs, enabling models to capture complex structures.

A number of works embed topology information into graph nodes by designing tailored attention mechanisms or positional and structural encodings without message-passing GNNs. 
GT~\citep{GT} is the pioneering work of GTs by integrating Laplacian positional encoding. 
In the following years, a series of works spring up.
SAN~\citep{SAN} incorporates both sparse and global attention mechanisms in each layer, utilizing Laplacian positional encodings for the nodes. 
Graphormer~\citep{Graphormer} attains state-of-the-art performance in graph-level prediction tasks with centrality encoding, spatial encoding and edge encoding. 
EGT~\citep{EGT} introduces the edge channel into attention mechanisms and adopts an SVD-based positional encoding instead of Laplacian positional encoding.
While self-attention is commonly constrained by quadratic complexity, our proposed GEA demonstrates a linear computational complexity with respect to both the number of nodes and edges.

Furthermore, some works introduce hybrid architectures incorporating message-passing GNNs. 
For instance, GraphTrans~\citep{GraphTrans} utilizes a stack of GNN layers before establishing full connectivity attention within the graph. 
Focusing on kernel methods, SAT~\citep{SAT} introduces a structure-aware attention mechanism using GNNs to extract a subgraph representation rooted at each node before computing the attention. 
A recent breakthrough emerged with the introduction of GraphGPS~\citep{GPS}, the first parallel framework that combines local message-passing and a global attention mechanism with various positional and structural encodings.
While these methods have achieved competitive performance, they overlook external information in the graph.
Therefore, to alleviate this issue, we design GEAET, which inherits the merits of the GEA network, message-passing GNN and Transformer, leverages inter-graph correlations, local structure and global interaction information.

\section{Method}
In the following, we denote a graph as $\graph = (\Vcal, \Ecal)$, where $\Vcal$ represents the set of nodes and $\Ecal$ represents the edges. 
The graph has $n=|\Vcal|$ nodes and $m=|\Ecal|$ edges. 
We denote the node features for a node $i\in\Vcal$ as $x_i$ and the features for an edge between nodes $i$ and $j$ as $e_{i,j}$. 
All node features and edge features are stored in matrices $\X\in\real^{n\times d}$ and $\E\in\real^{m\times d}$, respectively.
\subsection{Graph External Attention}
\label{sec:GEA}
We first revisit the Transformer, as illustrated in Figure~\ref{fig:self_attn}.
Transformer consists of two blocks: a self-attention module and a feed-forward network (FFN). 
Specifically, self-attention regards the graph as a fully connected graph and computes the attention of each node to every other node. 
With the input node features $\X$, self-attention linearly projects the input into 3 matrices: a query matrix ($\Q$), a key matrix ($\K$) and a value matrix ($\V$), where 
$\Q=\X \mathbf{W}_Q$, $\K=\X\mathbf{W}_K$ and $\V=\X \mathbf{W}_V$.
Then self-attention can be formulated as:
\begin{equation}\label{eq:Self_A}
    \begin{aligned}
    \A_{Self} =\softmax(\frac{\Q\K^T}{\sqrt{d_{out}}})\in\real^{n\times n}, \\
    \text{Self-Attn}(\X) =\A_{Self}\V\in\real^{n\times d_{out}}, 
    \end{aligned}
\end{equation}
where $\mathbf{W}_Q, \mathbf{W}_K, \mathbf{W}_V$ are trainable parameters and $d_{out}$ denotes the dimension of $\Q$.
The output of the self-attention is followed by both a skip connection and a FFN.

Self-attention on a graph can be viewed as employing a linear combination of node features within a single graph to refine node features. 
However, it exclusively focuses on the correlations among nodes within a single graph, overlooking implicit connections between nodes in different graphs, which may potentially limit its capacity and adaptability. 
\begin{figure*}[ht]   
    \newcommand{\wth}{1.}
    \centering
    {
    \includegraphics[width=\wth\textwidth]{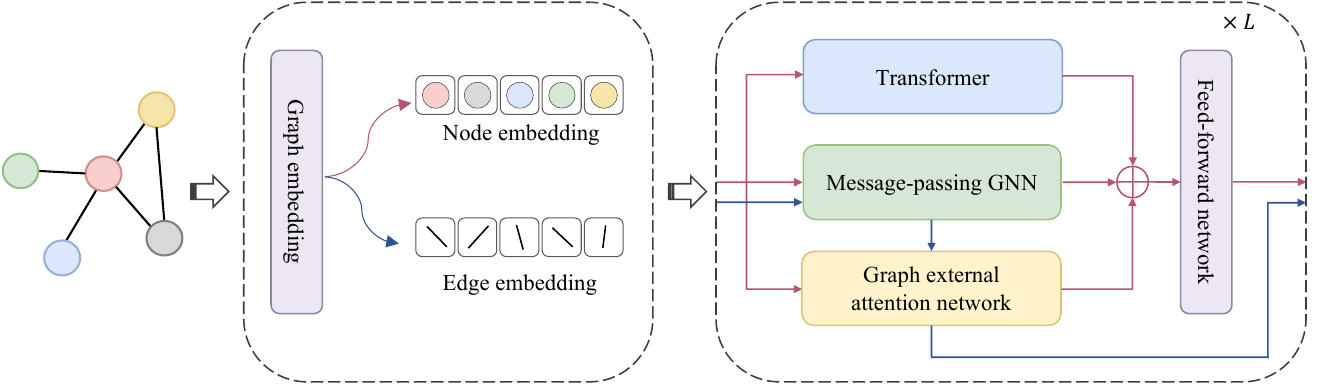}
    }
    \caption{Overall architecture of GEAET. It consists of a graph embedding layer and $L$ feature extraction layers.
    The graph embedding layer transforms graph data into node embeddings $\X$ and edge embeddings $\E$. It computes positional encodings, which are added to the node embeddings as inputs to the feature extraction layers. 
    Each feature extraction layer consists of a graph external attention network, a message-passing GNN and a Transformer to extract inter-graph correlations, local structures and global interaction information. 
    Finally, this information is integrated using a feed-forward network (FFN) and then employed on the output embeddings for various graph tasks.
    }
    \label{fig:GEAET_architecture}
\end{figure*}

Thus, inspired by~\citep{external_attn}, we introduce a novel method called GEA, as shown in Figure~\ref{fig:GEAttn}. 
It calculates attention between the node features of the input graph and the external units, using:
\begin{equation}\label{eq:Ge_A}
    \begin{aligned}
    \A_{GE} =\norm({\X\U^T})\in\real^{n\times S}, \\
    \text{GE-Attn}(\X) =\A_{GE}\U\in\real^{n\times d}, 
    \end{aligned}
\end{equation} 
where $\U\in\real^{S\times d}$ is a learnable parameter independent of the input graph, 
it can be viewed as an external unit with $S$ nodes, serving as shared memory for all input graphs.
In self-attention, $\A_{Self}$ represents the similarities between the nodes of the input graph, 
while in GEA, $\A_{GE}$ denotes the similarities between the nodes of the input graph and the external unit.
Considering the sensitivity of the attention matrix to the scale of input features,  
we apply a double-normalization technique~\citep{guo2021pct} on $\A_{GE}$.
The double-normalization process normalizes both columns and rows separately, it can be expressed as:
\begin{equation}\label{eq:d_norm}
    \begin{aligned}
        \tilde{\alpha}_{i,j} &=({\X\U^T})_{i,j}, \\
        \hat{\alpha}_{i,j} &=\exp(\tilde{\alpha}_{i,j})/\sum\nolimits_{k=0}^n\exp(\tilde{\alpha}_{k,j}), \\
        {\alpha}_{i,j} &=\hat{\alpha}_{i,j}/\sum\nolimits_{k=0}^S\hat{\alpha}_{i,k}.
    \end{aligned}
\end{equation}
In practical applications, for boosting the capability of the network, we utilize two distinct external units for the key and value. 
Furthermore, to leverage the edge information within the input graph, we employ additional external units for edge features and a shared unit to store the connections between edges and nodes:
\begin{equation}\label{eq:Ge_out}
    \begin{aligned}
    \X_{out} &=\norm({\X\U_{s}\U_{nk}^T})\U_{nv},\\
    \E_{out} &=\norm({\E\U_{s}\U_{ek}^T})\U_{ev},
    \end{aligned}
\end{equation}
where $\U_{s}\in\real^{d\times d}$ is a shared unit to store the connections between edges and nodes; $\U_{nk}, \U_{nv} \in\real^{S\times d}$ are external key-value units for nodes, while $\U_{ek}, \U_{ev} \in\real^{S\times d}$ are external key-value units for edges.

Within the Transformer architecture, self-attention is computed across various input channels in multiple instances, a technique referred to as multi-head attention. 
Multi-head attention can capture diverse node relations, enhancing the ability of the attention mechanism. 
Similarly, take nodes as an example, 
the relations between nodes within the graph and external units are various. 
Therefore, we adopt an analogous approach, it can be written as:
\begin{equation}\label{eq:Ge_out}
    \begin{aligned}
        h_{i} &=\text{GE-Attn}(\X_i,\U_{nk},\U_{nv}),\\
        \X_{out} &=\text{MultiHeadGEA}(\X,\U_{nk},\U_{nv})\\
        &=\Concat(h_1,...,h_H)\W_o,
    \end{aligned}
\end{equation}
where $h_i$ represents the $i$-th head, $H$ is the total number of heads, $\W_o$ is a linear transformation matrix, 
$\U_{nk},\U_{nv}\in\real^{S\times d}$ serve as shared memory units for different heads.
Finally, the output of the GEA is followed by a skip connection forming a Graph External Attention Network (GEANet).
\begin{table*}[ht]
    \centering
    \caption{Comparison of GEAET with baselines on 6 datasets. Best results are colored: \first{first}, \second{second}, \third{third}.}
    \label{tab:sota}
    \fontsize{8.25pt}{8.25pt}\selectfont
    \scalebox{0.9}{
    \begin{tabular}{p{4cm}cccccc}
    \toprule
         {\bf Model} & {\bf CIFAR10} & {\bf MNIST} & {\bf PATTERN} & {\bf Peptides-Struct} & {\bf PascalVOC-SP} & {\bf COCO-SP} \\
          & {Accuracy(\%) $\uparrow$} & {Accuracy(\%) $\uparrow$} & {Accuracy(\%) $\uparrow$} & {MAE $\downarrow$} & {F1 score $\uparrow$} & {F1 score $\uparrow$}\\
    \midrule
    GCN {\tiny\cite{GCN2017semi}} & 55.710 $\pm$ 0.381 & 90.705 $\pm$ 0.218 & 71.892 $\pm$ 0.334 &  0.3496 $\pm$ 0.0013 &  0.1268 $\pm$ 0.0060 &  0.0841 $\pm$ 0.0010 \\
    GINE {\tiny\cite{GIN}} & -- & -- & -- & 0.3547 $\pm$ 0.0045 & 0.1265 $\pm$ 0.0076 & 0.1339 $\pm$ 0.0044 \\
    GIN {\tiny\cite{GIN}} & 55.255 $\pm$ 1.527 & 96.485 $\pm$ 0.252 & 85.387 $\pm$ 0.136 & -- & -- & -- \\
    GAT {\tiny\cite{GAT}} & 64.223 $\pm$ 0.455 & 95.535 $\pm$ 0.205 & 78.271 $\pm$ 0.186 & -- & -- & -- \\
    GatedGCN {\tiny\cite{GatedGCN}} & 67.312 $\pm$ 0.311 & 97.340 $\pm$ 0.143 & 85.568 $\pm$ 0.088 & 0.3357 $\pm$ 0.0006 & 0.2873 $\pm$ 0.0219 & 0.2641 $\pm$ 0.0045 \\
    PNA {\tiny\cite{PNA}} & 70.350 $\pm$ 0.630 & 97.940 $\pm$ 0.120 & -- & -- & -- & --\\
    DGN {\tiny\cite{DGN}} & 72.838 $\pm$ 0.417 & -- & 86.680 $\pm$ 0.034 & -- & -- & --   \\
    DRew   {\tiny\cite{Drew}} &  -- & -- &  --  & 0.2536 $\pm$ 0.0015 & 0.3314 $\pm$ 0.0024 &  --  \\
    \midrule
    CRaWl~{\tiny\cite{CRaWl}} & 69.013 $\pm$ 0.259 &  97.944 $\pm$ 0.050 & --  & -- & -- & -- \\
    GIN-AK+~{\tiny\cite{GIN_AK}} & 72.190 $\pm$ 0.130 & -- & 86.850 $\pm$ 0.057  & -- & -- & -- \\
    \midrule
    SAN {\tiny \cite{SAN}} & -- & -- & 86.581 $\pm$ 0.037 & 0.2545 $\pm$ 0.0012 & 0.3230 $\pm$ 0.0039 & 0.2592 $\pm$ 0.0158\\
    K-Subgraph SAT {\tiny \cite{SAT}} & -- & -- &  86.848 $\pm$ 0.037 & -- & -- & -- \\
    EGT {\tiny \cite{EGT}} & 68.702 $\pm$ 0.409 & 98.173 $\pm$ 0.087 & 86.821 $\pm$ 0.020 &  -- &  -- &  -- \\
    GraphGPS {\tiny\cite{GPS}} &  72.298 $\pm$ 0.356 & 98.051 $\pm$ 0.126 & 86.685 $\pm$ 0.059 & 0.2500 $\pm$ 0.0005 & \third{0.3748 $\pm$ 0.0109} & \third{0.3412 $\pm$ 0.0044} \\
    LGI-GT   {\tiny\cite{LGI_GT}} &  -- & -- &  \third{86.930 $\pm$ 0.040} & -- & -- & -- \\
    GPTrans-Nano   {\tiny\cite{Drew}} &  -- & -- & 86.731 $\pm$ 0.085 &  --  &  -- &  --\\
    Graph-ViT/MLPMixer  {\tiny\cite{Graph_ViT}} &  73.960 $\pm$ 0.330 & \second{98.460 $\pm$ 0.090} &  -- & \second{0.2449 $\pm$ 0.0016} &  -- &  -- \\
    GRIT {\tiny\cite{Grit}} &  \second{76.468 $\pm$ 0.881} & 98.108 $\pm$ 0.111 &  \first{87.196 $\pm$ 0.076} & \third{0.2460 $\pm$ 0.0012} &  -- &  --\\
    Exphormer {\tiny\cite{Exphormer}} &  \third{74.754 $\pm$ 0.194} & \third{98.414 $\pm$ 0.038} &  86.734 $\pm$ 0.008 &  0.2481 $\pm$ 0.0007 &  \second{0.3966 $\pm$ 0.0027} &  \second{0.3430 $\pm$ 0.0008} \\
    \midrule
    GEAET (ours) & \first{76.634 $\pm$ 0.427} & \first{98.513 $\pm$ 0.086}  & \second{86.993 $\pm$ 0.026} & \first{0.2445 $\pm$ 0.0013}& \first{0.4585 $\pm$ 0.0087}& \first{0.3895 $\pm$ 0.0050}\\
     \bottomrule
    \end{tabular}
    }
\end{table*}
\subsection{Graph External Attention Enhanced Transformer}
Figure~\ref{fig:GEAET_architecture} illustrates an overview of the proposed GEAET framework. 
GEAET consists of two components: graph embedding and feature extraction layers.
\paragraph{Graph Embedding.} 
For each input graph, we initially perform a linear projection of the input node features $\alpha_i\in\real^{d_{\alpha}}$ and edge features $\beta_{i,j}\in\real^{d_{\beta}}$, resulting in $d$-dimensional hidden features:
\begin{equation}\label{eq:Ge_out}
    \begin{aligned}
        \tilde{x_i^0} &=\W_x^0\alpha_i+u^0\in\real^{d},\\
        e_{ij}^0&=\W_e^0\beta_{i,j}+v^0\in\real^{d},
    \end{aligned}
\end{equation}
where $\W_x^0\in\real^{d \times d_{\alpha}}$, $\W_e^0\in\real^{d \times d_{\beta}}$ and $u^0,v^0\in\real^{d}$ are learnable parameters. 
Then, we use positional encoding to enhance the input node features:
\begin{equation}\label{eq:Ge_out}
    x_i^0 =\T^0p_i+\tilde{x_i^0},
\end{equation}
where $\T^0\in\real^{d \times k}$ is a learnable matrix and $p_i\in\real^{k}$ is positional encoding. 
It is noteworthy that the advantages of different positional encodings are dependent on the dataset.
\paragraph{Feature Extraction Layer.} 
At each layer, external feature information is captured by the GEANet and then aggregated with intra-graph information to update node features. 
The intra-graph information is obtained through a combination of message-passing GNN and Transformer. 
This process can be formulated as:
\begin{equation}\label{eq:Ge_out}
    \begin{aligned} 
        \X_M^{l+1},\E_M^{l+1} &={\MPNN}^l(\X^{l},\E^{l},\A),\\
        \X_{T}^{l+1} &={\TLayer}^l(\X^{l}),\\
        \X_{G}^{l+1},\E_{G}^{l+1} &={\GEANet}^l(\X^{l},\E_M^{l+1}),\\
    \end{aligned}
\end{equation}where ${\GEANet}$ refers to the graph external attention network introduced in Section~\ref{sec:GEA}, 
$\TLayer$ represents the Transformer layer with self-attention,  
$\A\in\real^{n\times n}$ is the adjacency matrix, 
${\MPNN}$ is an instance of a message-passing GNN to update node and edge representations as follows:
\begin{equation}\label{eq:Ge_out}
    \begin{aligned}
        x_{i}^{l+1} &=f_{node}(x_{i}^l,\{x_j^l \mid j\in\Ncal(i)\},e_{i,j}^l),\\
        e_{i,j}^{l+1} &=f_{edge}(x_{i}^l,x_{j}^l,e_{i,j}^{l}),\\
    \end{aligned}
\end{equation}
where $x_i^{l+1},x_i^{l},e_{i,j}^{l+1},e_{i,j}^{l}\in\real^d$, $l$ is the layer index, $i,j$ denotes the node index, $\Ncal(i)$ is the neighborhood of the $i$-th node  
and the functions $f_{node}$ and $f_{edge}$ with learnable parameters define any arbitrary message-passing GNN architecture~\citep{GCN2017semi,GatedGCN,GraphSage,GAT,GIN,GINE}. 

Finally, we employ an FFN block to aggregate node information to obtain the node representations $\X^{l+1}$.
Additionally, we employ $\E_{G}^{l+1}$ as the edge features for the $l+1$-th layer:
\begin{equation}\label{eq:Ge_out}
    \begin{aligned} 
        \X^{l+1} &=\FFN^l(\X_{G}^{l+1}+\X_{T}^{l+1}+\X_M^{l+1}),\\
        \E^{l+1} &=\E_{G}^{l+1},
    \end{aligned}
\end{equation}
where $\X_{T}^{l+1},\X_M^{l+1}\in\real^{n\times d}$ are the outputs of $l$-layer Transformer and message-passing GNN, $\X_{G}^{l+1},\E_{G}^{l+1}\in\real^{n\times d}$ are the outputs of $l$-layer GEANet.

See Appendix~\ref{sec:complexity} for the complexity analysis of GEAET.

\begin{table*}[ht]
    \centering
    \caption{Comparison of the classic message-passing GNN baselines with their variants augmented by GEANet on 5 benchmarks.}
    \label{tab:GNN_GEA1}
    \fontsize{8.25pt}{8.25pt}\selectfont
    \setlength\tabcolsep{6.25pt} 
    \scalebox{0.9}{
    \begin{tabular}{p{1.5cm}ccccc} 
    \toprule
         {\bf Model} & {\bf PascalVOC-SP} & {\bf COCO-SP} & {\bf Peptides-Struct} & {\bf Peptides-Func} & {\bf PCQM-Contact}  \\
              & {F1 score $\uparrow$} & {F1 score $\uparrow$} & {MAE $\downarrow$} & {AP $\uparrow$} & {MRR $\uparrow$} \\
    \midrule
    GCN       & 0.1268 $\pm$ 0.0060 & 0.0841 $\pm$ 0.0010 & 0.3496 $\pm$ 0.0013 &  0.5930 $\pm$ 0.0023 &  0.3234 $\pm$ 0.0006 \\
    +GEANet      & \textbf{0.2250 $\pm$ 0.0103} & \textbf{0.2096 $\pm$ 0.0041} & \textbf{0.2512 $\pm$ 0.0003} &  \textbf{0.6722 $\pm$ 0.0065} &  \textbf{0.3244 $\pm$ 0.0007} \\
    \midrule

    GINE      & 0.1265 $\pm$ 0.0076 & 0.1339 $\pm$ 0.0044 & 0.3547 $\pm$ 0.0045 &  0.5498 $\pm$ 0.0079 &  0.3180 $\pm$ 0.0027 \\
    +GEANet      & \textbf{0.2742 $\pm$ 0.0032} & \textbf{0.2410 $\pm$ 0.0028} & \textbf{0.2544 $\pm$ 0.0012} &  \textbf{0.6509 $\pm$ 0.0021} &  \textbf{0.3276 $\pm$ 0.0012} \\

    \midrule
    GatedGCN  & 0.2873 $\pm$ 0.0219 & 0.2641 $\pm$ 0.0045 & 0.3420 $\pm$ 0.0013 &  0.5864 $\pm$ 0.0077 &  0.3242 $\pm$ 0.0008 \\
    +GEANet      & \textbf{0.3933 $\pm$ 0.0027} & \textbf{0.3219 $\pm$ 0.0052} & \textbf{0.2547 $\pm$ 0.0009} &  \textbf{0.6485 $\pm$ 0.0035} &  \textbf{0.3321 $\pm$ 0.0008} \\

    \bottomrule
    \end{tabular}
    }
\end{table*}
\begin{table*}[t]
    \centering
    \caption{Comparison of the classic message-passing GNN baselines with their variants augmented by GEANet on 5 benchmarks.}
    \label{tab:GNN_GEA2}
    \fontsize{8.25pt}{8.25pt}\selectfont
    \setlength\tabcolsep{6.25pt} 
    \scalebox{0.9}{
    \begin{tabular}{p{1.5cm}ccccc} 
    \toprule
         {\bf Model} & {\bf PATTERN} & {\bf CLUSTER} & {\bf MNIST} & {\bf CIFAR10} & {\bf ZINC}  \\
              & {Accuracy(\%) $\uparrow$} & {Accuracy(\%) $\uparrow$} & {Accuracy(\%) $\uparrow$} & {Accuracy(\%) $\uparrow$} & {MAE $\downarrow$} \\
    \midrule
    GCN       & 71.892 $\pm$ 0.334 & 68.498 $\pm$ 0.976 & 90.705 $\pm$ 0.218 &  55.710 $\pm$ 0.381 &  0.367 $\pm$ 0.011 \\
    +GEANet      & \textbf{85.323 $\pm$ 0.128} & \textbf{74.015 $\pm$ 0.124} & \textbf{96.465 $\pm$ 0.054} &  \textbf{61.925 $\pm$ 0.271} &  \textbf{0.240 $\pm$ 0.008} \\
    \midrule

    GIN      & 85.387 $\pm$ 0.136 & 64.716 $\pm$ 1.553 & 96.485 $\pm$ 0.252 &  55.255 $\pm$ 1.527 &  0.526 $\pm$ 0.051 \\
    +GEANet      & \textbf{85.527 $\pm$ 0.015} & \textbf{66.370 $\pm$ 2.145} & \textbf{96.845 $\pm$ 0.097} &  \textbf{62.320 $\pm$ 0.221} &  \textbf{0.193 $\pm$ 0.001} \\

    \midrule
    GatedGCN  & 85.568 $\pm$ 0.088 & 73.840 $\pm$ 0.326 & 97.340 $\pm$ 0.143 &  67.312 $\pm$ 0.311 &  0.282 $\pm$ 0.015 \\
    +GEANet      & \textbf{85.607 $\pm$ 0.038} & \textbf{77.013 $\pm$ 0.224} & \textbf{98.315 $\pm$ 0.097} &  \textbf{73.857 $\pm$ 0.306} &  \textbf{0.218 $\pm$ 0.011} \\

    \bottomrule
    \end{tabular}
    }
\end{table*}

\section{Experiments} \label{sec:experiments}
In this section, we evaluate the empirical performance of GEANet and GEAET on a variety of graph datasets with graph prediction and node prediction tasks, 
including CIFAR10, MNIST, PATTERN, CLUSTER and ZINC from Benchmarking GNNs~\citep{benchmarkingGNN}, as well as PascalVOC-SP, COCO-SP, Petides-Struct, Petides-Func and PCQM-Contact from Long Range Graph Benchmark (LRGB;~\citealp{Lrgb}), and the TreeNeighbourMatch dataset~\citep{TreeNeighbourMatch}.
Detailed information is provided in the Appendix~\ref{sec:datasetdesc}.

We first compare our main architecture, GEAET, with the latest state-of-the-art models.
In addition, we integrate GEANet with the message-passing GNNs and compare it with the corresponding network to demonstrate the role of GEANet. 
Furthermore, we conduct a series of comparative experiments with Transformer, including visualization experiments on attention in molecular graphs, experiments varying the number of attention heads and positional encoding experiments.
Finally, we conduct ablation studies on each component of GEANet, including the external node unit, the external edge unit and the shared unit, to confirm the effectiveness of each component. 
More details on the experimental setup and hyperparameters are provided in the Appendix~\ref{sec:hyperparameter}, additional results are given in the Appendix~\ref{sec:additional_results}.

In summary, our experiments reveal that 
(a) GEAET architecture outperforms existing state-of-the-art methods on various datasets, 
(b) GEANet can be seamlessly integrated with some basic GNNs, significantly enhancing the performance, 
(c) GEANet shows better interpretability than Transformer and 
(d) GEANet is less dependent on positional encoding.

\subsection{Comparison with SOTAs}\label{sec:attentioncomp}

We compare our methods with several recent SOTA graph Transformers, including Exphormer, GRIT, Graph-ViT/MLPMixer, and numerous popular graph representation learning models, such as well-known message-passing GNNs (GCN, GIN, GINE, GAT, GatedGCN, PNA), and graph Transformers (SAN, SAT, EGT, GraphGPS, LGI-GT, GPTrans-Nano).
Additionally, we consider other recent methods with SOTA performance, such as DGN, DRew, CRaW1 and GIN-AK+.

As shown in Table~\ref{tab:sota}, for the 3 tasks from Benchmarking GNNs~\citep{benchmarkingGNN}, we observe that our GEAET achieves SOTA results on CIFAR10 and MNIST and ranks second on the PATTERN dataset.
For the 3 tasks on LRGB~\citep{Lrgb}, GEAET achieves the best results. 
It is noteworthy that the GEAET achieves F1 scores of 0.4585 on PascalVOC-SP and 0.3895 on COCO-SP, 
surpassing other models by a significant gap.
\subsection{Comparison with GNNs} \label{sec:baselines}
To clearly demonstrate the performance improvement of GEANet on graph representation learning models, we integrate GEANet with some commonly used message-passing GNNs, providing the models with the ability to learn graph external information. 
In our comparison, we evaluate our approach against the corresponding GNNs, with GNN results sourced from~\citet{benchmarkingGNN} or~\citet{Lrgb}. 
To maintain a fair comparison, our trained models strictly adhere to the parameter constraints without incorporating any positional encoding, consistent with~\citet{benchmarkingGNN} and~\citet{Lrgb}.
As depicted in Table~\ref{tab:GNN_GEA1} and Table~\ref{tab:GNN_GEA2}, GEANet significantly improves the performance of all base message-passing GNNs, including GCN~\citep{GCN2017semi}, GatedGCN~\citep{GatedGCN}, GIN~\citep{GIN} and GINE~\citep{GINE}, on various datasets simply by combining the output of GEANet with the output of the GNN.
This is achieved without any additional modifications, validating that GEANet can effectively alleviate issues in message-passing GNNs.

\subsection{Comparison with Self-Attention} \label{sec:baselines}
\begin{figure}[ht]
    \includegraphics[width=.25\columnwidth]{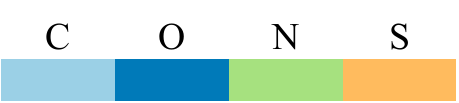} \\
    \begin{center}
    \includegraphics[width=.99\columnwidth]{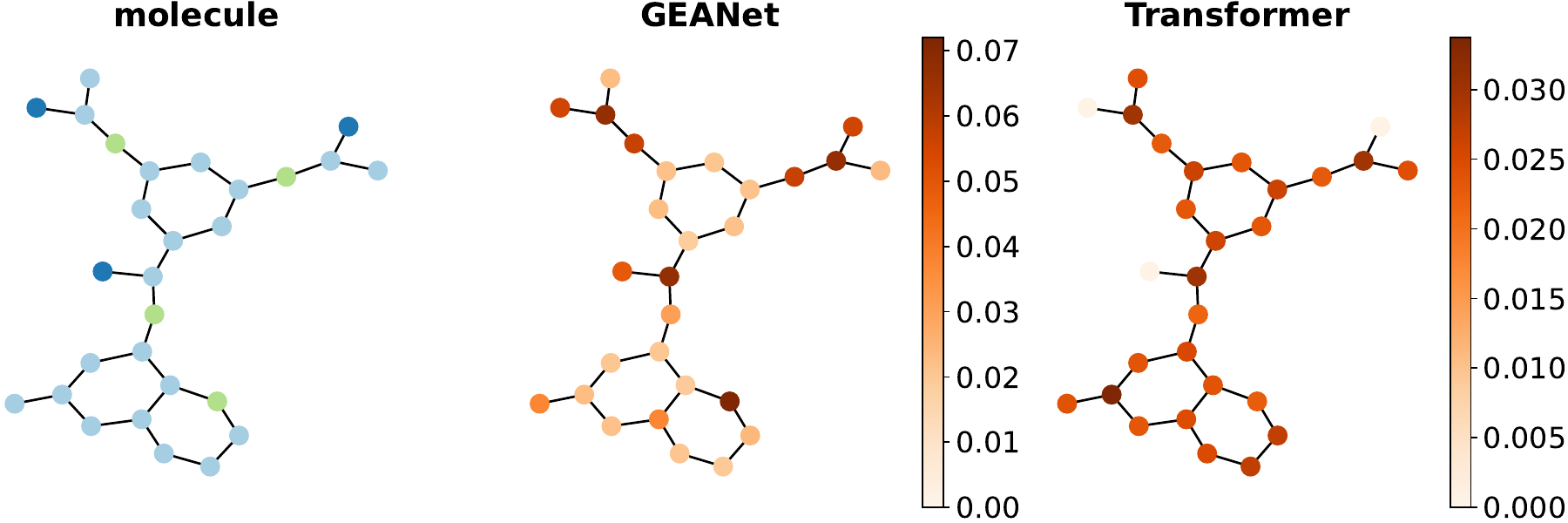}\\
    \includegraphics[width=.99\columnwidth]{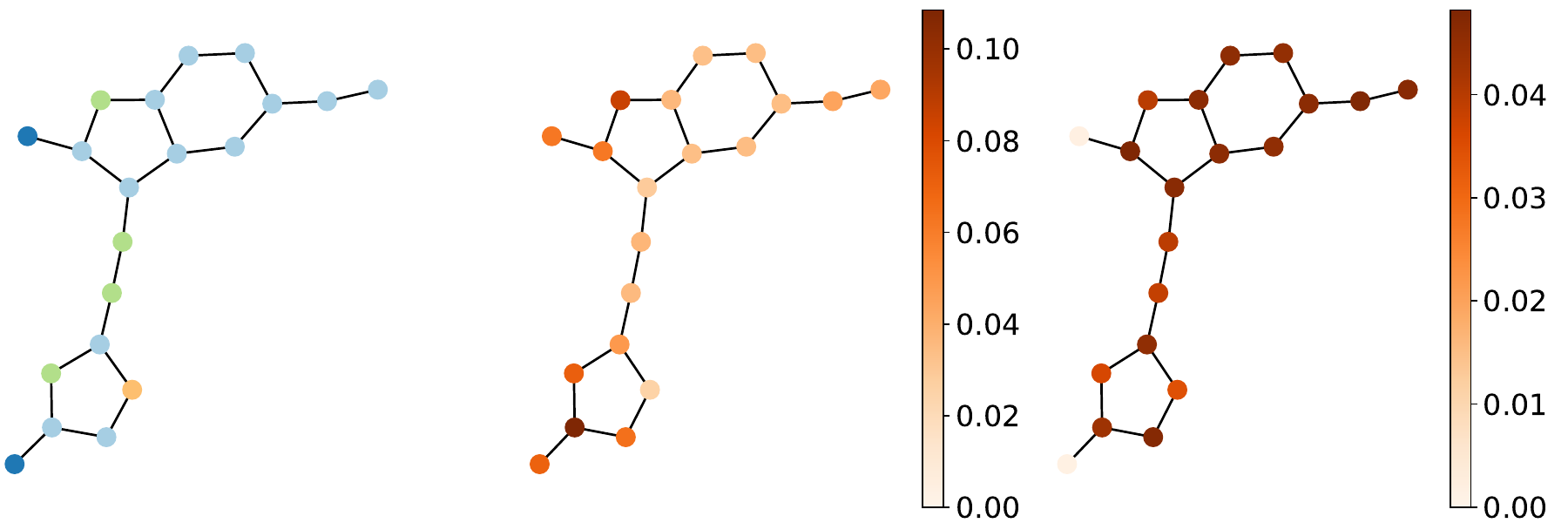}
    \caption{Attention visualization of GEANet and Transformer on ZINC molecular graphs. The left column shows two original molecular graphs, while the middle and right columns show the visualization results of attention scores with GEANet and Transformer, respectively.}
    \label{fig:zinc_attn_visual}
    \end{center}
\end{figure}

\paragraph{Attention Interpretation.} 
To better explain the attention mechanism, we respectively train a GEANet and a Transformer on the ZINC dataset and visualize the attention scores in Figure~\ref{fig:zinc_attn_visual}. The salient difference between the two models is that GEANet can capture the correlation between graphs, and thus we can attribute the following interpretability gains to that. While both models manage to identify some hydrophilic structures or functional groups, the attention scores learned by GEANet are sparser and more informative. GEANet focuses more on important atoms such as N and O, as well as atoms that connect different motifs. The attention distribution of GEANet is similar to the structural distribution of the original molecular graphs, which promotes to predict the restricted solubility more accurately. In contrast, Transformer does not utilize inter-graph correlations, resulting in poorer predictive performance.
More results are provided in the Appendix~\ref{sec:appendix_interpretation}.

\begin{figure}[ht]
        \pgfplotsset{
        every x axis/.append style={x axis line style=->},
        every non boxed y axis/.append style={y axis line style=->}
        }
    \centering
    \begin{tikzpicture}[scale=0.9]
      \pgfplotsset{
        every axis/.append style = {
          mark size = 0.75pt, 
        }
      }
      \begin{axis}[%
        axis x line*      = bottom,
        axis y line*      = left,
        axis line style   = ->,
        xlabel            = {Number of attention heads},
        ylabel            = {MAE},
        ticklabel style={font=\small},
        xtick             = {0,2,4,6,8,10},
        xticklabels       = {0,2,4,6,8,10},
        ytick             = {0.245,0.25,0.255,0.26,0.265},
        yticklabels       = {0.245,0.25,0.255,0.26,0.265},
        label style       = {font={\normalsize}},
        legend pos        = {north east},
        legend cell align = left,
        legend style      = {%
          draw = none,
        enlarge x limits  = 0.05,
        enlarge y limits  = 0.05
        },
        legend image post style = {
          scale = 1.0
        },
        legend columns=1, 
        legend style={
            font={\normalsize},
            /tikz/column 2/.style={
                column sep=5pt,
            },
        },
      ]
          \addplot[bar2, mark=*, line width = 0.75 pt, error bars/.cd, y fixed, y dir=both, y explicit] table[col sep = comma, x index = 0, y index = 3, y error index = 4]{data/heads.csv};
          \addplot[bar1, mark=*, line width = 0.75 pt, error bars/.cd, y fixed, y dir=both, y explicit] table[col sep = comma, x index = 0, y index = 1, y error index = 2] {data/heads.csv};

          \legend{GCN+GEANet,GCN+Transformer};
      \end{axis}
  \end{tikzpicture}
     \caption{Test MAE with different number of attention heads.}
    \label{fig:effect_of_head}
\end{figure}
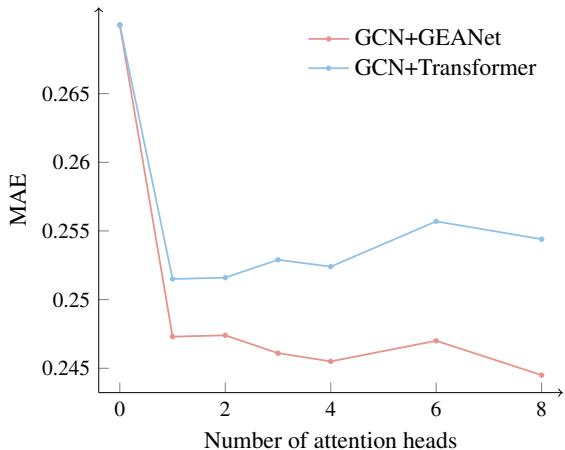

\paragraph{Impact of Attention Heads.} 
We investigate the impact on the number of attention heads with two attention mechanisms. 
Figure~\ref{fig:effect_of_head} shows the MAE values with either GCN + Transformer or GCN + GEANet on the Peptides-Struct dataset. 
For a fair comparison, both models use Laplacian positional encoding, with the same number of layers and approximately total number of parameters (about 500k). 
Heads = 0 corresponds to a pure GCN network without attention. 
The introduction of attention mechanism significantly improves performance.
GEANet achieves the best performance with 8 heads. 
In contrast, Transformer performs best with one head, suggesting that multiple self-attention heads do not enhance performance. 
Notably, GEANet consistently outperforms self-attention across various numbers of heads.

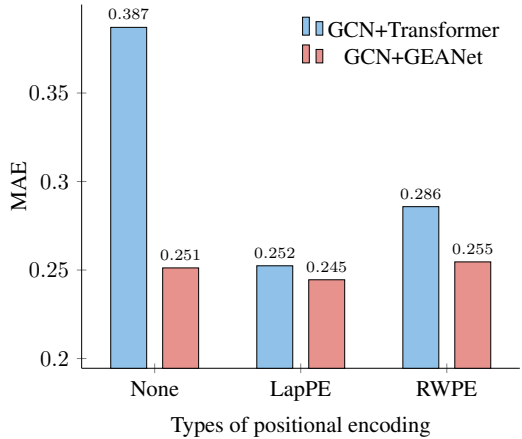
\begin{figure}[h]
        \begin{tikzpicture}[scale=.85]
     
    \begin{axis} [ybar = .25cm,
        yshift            = 5mm,  
        ylabel shift = -2 mm,
        bar width = 16pt,
        axis lines = left,
        xmin = 0.5, 
        xmax = 3.5,
        ymax = 0.35,
        label style       = {font={\normalsize}},
        ticklabel style={font=\normalsize},
        ylabel={MAE},
        xlabel={Types of positional encoding},
        enlarge y limits = {abs = .05},
        ytick = {0.2,0.25,0.3,0.35},
        xtick = {1,2,3},
        xticklabels={\normalsize None, \normalsize LapPE,\normalsize RWPE},
        nodes near coords,
        every node near coord/.append style={font=\scriptsize},
        nodes near coords style={/pgf/number format/.cd,fixed, zerofill,precision=3},
        legend style={font=\normalsize, draw=none},
    ]
    \addplot[fill=bar1] coordinates {(1,0.3871)(2,0.2524)(3,0.2858)}; % self
    \addplot[fill=bar2] coordinates {(1,0.2512)(2,0.2445)(3,0.2546)}; % external

    \legend{\normalsize GCN+Transformer,GCN+GEANet,123}
    \end{axis}
    
    \end{tikzpicture}
    \caption{Test MAE with different positional encodings.}
    \label{fig:pos_enc}
\end{figure}

\paragraph{Impact of Positional Encoding.} 
We conduct an ablation study on the Peptides-Struct dataset to assess the impact of positional encoding on GEANet and Transformer.
Similar to the experiments with attention heads, we utilize a parallel architecture consisting of an GCN block and an attention block. 
Figure~\ref{fig:pos_enc} shows the MAE values obtained with and without positional encoding, including Random Walk Positional Encoding (RWPE;~\citealp{RWPEli,lspe}) and Laplacian Positional Encoding (LapPE;~\citealp{GT,SAN}). 
The Transformer with self-attention performs poorly without positional encoding. 
The utilization of LapPE and RWPE improves performance to some extent. 
In contrast, GEANet achieves an good performance without positional encoding. 
For GEANet, we observe that LapPE can enhance performance, while RWPE decreases performance. 
On the whole, GEANet is less dependent on positional encoding compared to Transformer.

\subsection{GEAET Mitigates Over-Squashing} \label{sec:baselines}
\begin{figure}[htb]
    \pgfplotsset{
        every x axis/.append style={x axis line style=->},
        every non boxed y axis/.append style={y axis line style=->}
        
        }
    \begin{tikzpicture}[scale=0.9]
      \pgfplotsset{
        every axis/.append style = {
          mark size = 0.75pt, 
        }
      }
      \begin{axis}[%
        grid,
        grid style={dashed, gray!30}, 
        axis x line*      = bottom,
        axis y line*      = left,
        axis line style   = {dashed, gray!30},
        xlabel            = {$r$},
        ylabel            = {Accuracy},
        ticklabel style={font=\small},
        xtick             = {2,3,4,5,6,7},
        xticklabels       = {2,3,4,5,6,7},
        ytick             = {0, 0.1, 0.2, ..., 1.0},
        yticklabels       = {0.0,0.1,0.2,0.3,0.4,0.5,0.6,0.7,0.8,0.9,1.0},
        label style       = {font={\normalsize}},
        legend pos        = {south east},
        legend cell align = left,
        legend style      = {%
          draw = none,
        enlarge x limits  = 0.0,
        enlarge y limits  = 0.0
        },
        legend image post style = {
          scale = 1
        },
        legend columns=1, 
        legend style={
            at={(1.1,0.45)}, 
            font={\normalsize},
        },
      ]
          \addplot[line1, mark=otimes*, mark size=1.5,line width=1pt] table[col sep=comma, x index=0, y index=1] {data/tree.csv};
          \addplot[line2, mark=diamond*, mark size=1.5,line width=1pt] table[col sep=comma, x index=0, y index=2] {data/tree.csv};
          \addplot[line3, mark=pentagon*, mark size=1.5,line width=1pt] table[col sep=comma, x index=0, y index=3] {data/tree.csv};
          \addplot[line4, mark=triangle*,mark size=1.5, line width=1pt] table[col sep=comma, x index=0, y index=4] {data/tree.csv};
          \addplot[line6, mark=star,mark size=1.5, line width=1pt] table[col sep=comma, x index=0, y index=5] {data/tree.csv};
          \addplot[line7, mark=square*, mark size=1.3,line width=1pt] table[col sep=comma, x index=0, y index=6] {data/tree.csv};

          \legend{GCN,GINE,GAT,GIN,GatedGCN,GEAET};
      \end{axis}
  \end{tikzpicture}
     \caption{Test accuracy with different problem radius $r$ on the TreeNeighbourMatch dataset.}
    \label{fig:tree}
\end{figure}
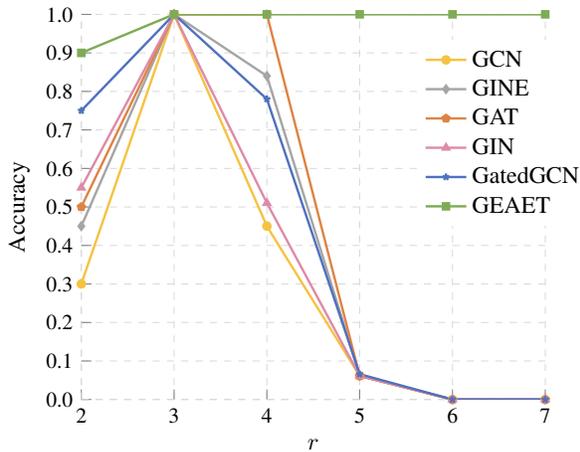

The TreeNeighbourMatch dataset~\citep{TreeNeighbourMatch} is used to provide an intuition of over-squashing.
On this dataset, each example is represented with a binary tree of depth $r$. 
Figure~\ref{fig:tree} shows the performance comparison of the standard message-passing GNNs with our proposed GEAET on the TreeNeighbourMatch dataset. 
The results indicate that our GEAET generalizes well on the dataset up to $r = 7$, effectively alleviating the issue of over-squashing. 
In contrast, the 5 GNN methods fail to generalize effectively from $r = 4$.
This is consistent with the perspective of~\citet{TreeNeighbourMatch} that GNN suffers from over-squashing due to a fixed-length embedding of the graph. 
Our method addresses this problem by utilizing graph external attention mechanisms to transmit long-range information.

\subsection{Ablation Experiments} \label{sec:baselines}
To assess the practicality of our model design choices, we conduct multiple ablation experiments on MNIST and PATTERN. 
The results are shown in Table~\ref{tab:ablation}. 
We notice that the removal of either external node units, external edge units, or external shared units all leads to poorer performance, which demonstrates the soundness of our architectural decisions.
\begin{table}[h]
    \centering
    \caption{Ablation study on MNIST and PATTERN datasets for GEA components. The metric is the Accuracy(\%).}
    \label{tab:ablation}
    \fontsize{8.25pt}{8.25pt}\selectfont
    \setlength\tabcolsep{6.25pt} 
    \scalebox{0.9}{
    \begin{tabular}{ccc|cc} 
    \toprule
         {\bf Node} & {\bf Edge} &{\bf Share} &{\bf MNIST} & {\bf PATTERN}  \\
    \midrule
    --          & --         & --           & 98.274 $\pm$ 0.011 & 86.882 $\pm$ 0.028\\
    \midrule

    \checkmark  & --         & --           & 98.474 $\pm$ 0.013 & 86.956 $\pm$ 0.038\\
    \checkmark  & \checkmark & --           & 98.488 $\pm$ 0.082 & 86.959 $\pm$ 0.024\\    
    \checkmark  & \checkmark & \checkmark   & \textbf{98.513 $\pm$ 0.086} & \textbf{86.993 $\pm$ 0.026}\\    
    \bottomrule
    \end{tabular}
    }
\end{table}
\section{Conclusion}
Observing that existing graph representation learning methods are confined to internal information, we argue for the importance of inter-graph correlations.
Drawing inspiration from the idea that graphs with similar structures ought to have analogous representations, 
we propose GEA, a novel lightweight yet effective attention mechanism to exploit inter-graph correlations.
On this basis, we introduce GEAET to exploit local structure and global interaction information.
GEAET achieves state-of-the-art performance on various graph datasets, highlighting the significance of inter-graph correlations.
Nevertheless, GEAET is not the final chapter of our work; future efforts will focus on reducing the high memory cost and time complexity, as well as addressing the lack of upper bounds on expressive power.

\section*{Acknowledgements}

This work is supported by the National Science and Technology Major Project (2020AAA0106102) 
and National Natural Science Foundation of China (No.62376142, U21A20473, 62272285). 
The authors would also like to thank Dr. Junbiao Cui for his insightful opinions during the rebuttal period.

\section*{Impact Statement}

% Authors are \textbf{required} to include a statement of the potential 
% broader impact of their work, including its ethical aspects and future 
% societal consequences. This statement should be in an unnumbered 
% section at the end of the paper (co-located with Acknowledgements -- 
% the two may appear in either order, but both must be before References), 
% and does not count toward the paper page limit. In many cases, where 
% the ethical impacts and expected societal implications are those that 
% are well established when advancing the field of Machine Learning, 
% substantial discussion is not required, and a simple statement such 
% as the following will suffice:

% ``
This paper presents work whose goal is to advance the field of 
Machine Learning. There are many potential societal consequences 
of our work, none which we feel must be specifically highlighted here.
% ''

% The above statement can be used verbatim in such cases, but we 
% encourage authors to think about whether there is content which does 
% warrant further discussion, as this statement will be apparent if the 
% paper is later flagged for ethics review.

\nocite{langley00}

\bibliographystyle{icml2024}

%%%%%%%%%%%%%%%%%%%%%%%%%%%%%%%%%%%%%%%%%%%%%%%%%%%%%%%%%%%%%%%%%%%%%%%%%%%%%%%
%%%%%%%%%%%%%%%%%%%%%%%%%%%%%%%%%%%%%%%%%%%%%%%%%%%%%%%%%%%%%%%%%%%%%%%%%%%%%%%
% APPENDIX
%%%%%%%%%%%%%%%%%%%%%%%%%%%%%%%%%%%%%%%%%%%%%%%%%%%%%%%%%%%%%%%%%%%%%%%%%%%%%%%
%%%%%%%%%%%%%%%%%%%%%%%%%%%%%%%%%%%%%%%%%%%%%%%%%%%%%%%%%%%%%%%%%%%%%%%%%%%%%%%

\newpage
\appendix
\onecolumn

\section{Dataset Descriptions} \label{sec:datasetdesc}

We evaluate our method on diverse datasets, including 5 graph benchmark datasets from Benchmarking GNNs, 5 long-range dependency graph datasets from LRGB and the TreeNeighbourMatch dataset. 
Below, we provide descriptions of the datasets and present summary statistics in Table~\ref{tb:datasets_info}.

\paragraph{MNIST and CIFAR10.}
MNIST and CIFAR10~\citep{benchmarkingGNN} represent the graphical counterparts of their respective image classification datasets of the same name. 
A graph is formed by creating an 8-nearest neighbor graph of the SLIC superpixels of the images. 
These datasets pose 10-class graph classification challenges. 
For MNIST, the resulting graphs have sizes ranging from 40 to 75 nodes, while for CIFAR10, the graphs vary between 85 and 150 nodes. 
The classification tasks involve the standard dataset splits of 55K/5K/10K for MNIST and 45K/5K/10K for CIFAR10, corresponding to train/validation/test graphs. 
These datasets serve as sanity checks, with an expectation that most GNNs would achieve close to 100\% accuracy for MNIST and satisfactory performance for CIFAR10. 

\paragraph{PATTERN and CLUSTER.}
PATTERN and CLUSTER~\citep{benchmarkingGNN} are synthetic datasets derived from the Stochastic Block Model (SBM;~\citealp{SBM}) on node classification tasks.
PATTERN is to determine whether a node belongs to one of the 100 predefined subgraph patterns, while CLUSTER is to classify nodes into 6 distinct clusters with identical distributions. The unique feature of PATTERN involves recognizing nodes belonging to randomly generated sub-graph patterns, while CLUSTER entails inferring the cluster ID for each node in graphs composed of 6 SBM-generated clusters. We use the splits as is used in~\citep{benchmarkingGNN}.

\paragraph{ZINC.}
ZINC~\citep{benchmarkingGNN} is a graph regression dataset derived from a subset of molecular graphs (12K out of 250K) sourced from a freely available database of commercially accessible compounds~\citep{ZINC}. 
The molecular graphs in ZINC range from 9 to 37 nodes, where each node represents a heavy atom (with 28 possible atom types) and each edge signifies a bond (with 3 possible types). 
The primary task is to regress a molecular property known as constrained solubility. 
The dataset includes a predefined train/validation/test split of 10K/1K/1K instances.

\paragraph{PascalVOC-SP and COCO-SP.}
PascalVOC-SP and COCO-SP~\citep{Lrgb} are graph-based versions of image datasets with larger images and involve the task of node classification, specifically the semantic segmentation of superpixels. 
These datasets respectively derived from the Pascal VOC 2011 image dataset~\citep{VOC} and the MS COCO image dataset through SLIC superpixelization, present a more intricate node classification challenge compared to CIFAR10 and MNIST. 
Each superpixel node is associated with a specific object class, making them node classification datasets with a focus on regions of images belonging to particular classes.

\paragraph{Peptides-Func and Peptides-Struct.} 
Peptides-Func and Peptides-Struct~\citep{Lrgb} are derived from 15,535 peptides with a total of 2.3 million nodes sourced from SAT-Pdb~\citep{peptides}. 
The graphs exhibit large sizes, averaging 150.94 nodes per graph and a mean graph diameter of 56.99. 
Specifically suited for benchmarking graph Transformers or expressive GNNs capable of capturing long-range interactions. Peptides-func involves multi-label graph classification into 10 nonexclusive peptide functional classes, while Peptides-struct focuses on graph-level regression predicting 11 3D structural properties of the peptides.

\paragraph{PCQM-Contact.} 
PCQM-Contact~\citep{Lrgb} is derived from PCQM4Mv2 and corresponding 3D molecular structures, where the task is a binary link prediction task. 
This dataset contains 529,434 graphs with a total of 15 million nodes, where each graph represents a molecular graph with explicit hydrogens. 
All graphs in PCQM-Contact are extracted from the PCQM4M training set, specifically those with available 3D structure and filtered to retain only those with at least one contact.

\paragraph{TreeNeighbourMatch.}
TreeNeighbourMatch is a synthetic dataset introduced by~\citet{TreeNeighbourMatch} to illustrate the challenge of over-squashing in GNNs. It features binary trees of controlled depth that simulate an exponentially-growing receptive field with a problem radius $r$. The task is to predict a label for the target node, situated in one of the leaf nodes, necessitating information propagation from all leaves to the target node. This setup exposes the issue of over-squashing at the target node due to the need for comprehensive long-range signal incorporation before label prediction.

\begin{table}[htp]
    \centering
    \caption{Summary statistics of datasets used in this study.}
    \fontsize{8.5pt}{8.5pt}\selectfont
    \setlength\tabcolsep{6.25pt} 
    \scalebox{1}{
    \begin{tabular}{lcccccc}
    \toprule
         {\bf Dataset} & {\bf Graphs} & {\bf Avg. nodes} & {\bf Avg. edges} & {\bf Prediction Level} & {\bf Task} &{\bf Metric} \\
    \midrule
    MNIST & 70,000&  70.6 & 564.5 & graph& 10-class classif. & Accuracy\\
    CIFAR10 & 60,000 & 117.6 & 941.1 &  graph & 10-class classif. & Accuracy\\
    PATTERN & 14,000 & 118.9 & 3,039.3  & inductive node & binary classif. & Accuracy\\
    CLUSTER & 12,000 & 117.2 & 2,150.9  & inductive node & 6-class classif. & Accuracy\\
    ZINC & 12,000 & 23.2 & 24.9 & graph & regression & MAE\\
    \midrule
    PascalVOC-SP & 11,355 & 479.4 & 2,710.5 & inductive node & 21-class classif. & F1\\
    COCO-SP & 123,286 & 476.9 & 2,693.7 & inductive node & 81-class classif. & F1\\
    PCQM-Contact & 529,434 & 30.1 & 61.0 & inductive link & link ranking & MRR\\
    Peptides-Func & 15,535 & 150.9 & 307.3 & graph & 10-class classif. & Avg. Precision\\
    Peptides-Struct & 15,535 & 150.9 & 307.3 & graph & regression & MAE\\
    \midrule
    TreeNeighbourMatch(r=2) & 96      & 7     & 6   & inductive node & 4-class classif. & Accuracy\\
    TreeNeighbourMatch(r=3) & 32,000  & 15    & 14  & inductive node & 8-class classif. & Accuracy \\
    TreeNeighbourMatch(r=4) & 64,000  & 31    & 30  & inductive node & 16-class classif. & Accuracy \\
    TreeNeighbourMatch(r=5) & 128,000 & 63    & 62  & inductive node & 32-class classif. & Accuracy \\
    TreeNeighbourMatch(r=6) & 256,000 & 127   & 126 & inductive node & 64-class classif. & Accuracy \\
    TreeNeighbourMatch(r=7) & 512,000 & 255   & 254 & inductive node & 128-class classif. & Accuracy \\
    \bottomrule
    \end{tabular}
    }
     \label{tb:datasets_info}
\end{table}
\section{Hyperparameter Choices and Reproducibility} \label{sec:hyperparameter}

\paragraph{Hyperparameter Choice.}
In our hyperparameter search, we attempt to adjust the number of heads in GEANet, as well as hyperparameters related to positional encoding, message-passing GNN type and Transformer. 
Considering the large number of hyperparameters and datasets, we do not conduct an exhaustive search or grid search. 
For a fair comparison, we follow commonly used parameter budgets: for benchmarking datasets from Benchmarking GNNs~\citep{benchmarkingGNN}, a maximum of 500k parameters for PATTERN and approximately 100k parameters for MNIST and CIFAR10; for datasets from LRGB~\citep{Lrgb}, we adhere to a parameter budget of 500k. 
See Table~\ref{tab:hyperparams_GEAET1} for detailed information. 
\paragraph{Optimization.}
We use the AdamW~\citep{adamw} optimizer in all our experiments, with the default settings of $\beta_1=0.9$, $\beta_2=0.999$ and $\epsilon= 10^{-8}$, and use a cosine scheduler~\citep{cosine}.
The choice of loss function, length of the warm-up period, base learning rate and total number of epochs are adjusted based on the dataset.

\begin{table}[ht]
\centering
\caption{Hyperparameters used for GEAET on 6 datasets.}
\label{tab:hyperparams_GEAET1}
\begin{tabular}{l|lllllll} 
\toprule
{\bf Hyperparameter}    & {\bf CIFAR10} & {\bf MNIST}   & {\bf PATTERN} & {\bf Peptides-Struct} & {\bf PascalVOC-SP}  & {\bf COCO-SP}\\ 
\hline
Layers          & 5         & 5          & 7           & 6        & 8          & 8   \\
Hidden Dim $d$       & 40        & 40         & 64          & 224      & 68         & 68   \\
MPNN                 & GatedGCN  & GatedGCN   & GatedGCN    & GCN      & GatedGCN   & GatedGCN   \\
Self Attention         & Transformer & Transformer  & Transformer   & None     & Transformer  & Transformer   \\
External Network    & GEANet       & GEANet        & GEANet         & GEANet      & GEANet        & GEANet    \\
Self Heads          & 4         & 4          & 4           & None     & 4          & 4   \\
External Heads      & 4         & 4          & 4           & 8        & 4          & 4   \\
Unit Size $S$      & 10         & 10          & 16           & 28        & 17          & 17   \\
\midrule
PE               & ESLapPE-8 & ESLapPE-8  & RWPE-16     & LapPE-10 & None       & None   \\ 
PE Dim              & 8         & 8          & 7           & 16       & None       & None   \\ 
\midrule
Batch Size          & 16        & 16         & 32          & 200      & 50         & 50   \\
Learning Rate       & 0.001     & 0.001      & 0.0005      & 0.001    & 0.001      & 0.001   \\
Num Epochs          & 150       & 150        & 100         & 250      & 200        & 200   \\ 
Warmup Epochs       & 5         & 5          & 5           & 5        & 10         & 10  \\ 
Weight Decay        & 1e-5      & 1e-5       & 1e-5        & 0        & 0          & 0   \\ 
\midrule
Num Parameters      & 113,235 & 113,155      & 429,052     & 463,211  & 506,213    & 505,661   \\
\bottomrule
\end{tabular}

\end{table}

\section{Additional Results} \label{sec:additional_results}
We provide additional results here, including detailed results from the attention heads and position encoding experiments, the results of GEAET on the link prediction task of the PCQM-Contact dataset and a substantial number of additional ablation studies. 
\paragraph{Impact of Attention Heads.}
We conduct experiments with different numbers of attention heads on the Peptides-Struct and Peptides-Func datasets. 
We adopt a framework where the base GNN and the attention block (Transformer or GEANet) operate in parallel, with the output of the model at each layer being the sum of the outputs from the GNN and the attention block. 
To ensure fairness, we use the same number of layers, apply Laplacian positional encoding (LapPE), use GCN as the base GNN and keep the total number of parameters to about 500k. 
The detailed results are shown in Table~\ref{tab:heads}, where all results are averaged over 4 different random seeds. 
We observe that having multiple attention heads does not significantly improve Transformer for both datasets. However, there is a notable improvement in GEANet, with the lowest MAE achieved with 8 heads on Peptides-Struct and the best AP achieved with 8 heads on Peptides-Func.
\begin{table*}[ht]
    \centering
    \caption{The results of Transformer and GEANet with different number of attention heads.}
    \label{tab:heads}
    \fontsize{8.25pt}{8.25pt}\selectfont
    \setlength\tabcolsep{6.25pt} 
    \scalebox{0.9}{
    \begin{tabular}{p{2.5cm}cccccc} 
    \toprule
         {\bf Model} & {\bf \#Layers} & {\bf Positional} & {\bf \#Heads} & {\bf \#Parameters} &{\bf Peptides-Struct } & {\bf Peptides-Func } \\
         &           &   {\bf    Encoding }  &        & &MAE $\downarrow$ &AP $\uparrow$   \\
    \midrule
    GCN + Transformer & 6 & LapPE &2 & 490,571 & 0.2516 $\pm$ 0.0031 &  0.6644 $\pm$ 0.0052\\
    GCN + Transformer & 6 & LapPE &3 & 490,571 & 0.2529 $\pm$ 0.0012 &  0.6688 $\pm$ 0.0072\\
    GCN + Transformer & 6 & LapPE &4 & 490,571 & 0.2524 $\pm$ 0.0017 &  0.6634 $\pm$ 0.0033\\
    GCN + Transformer & 6 & LapPE &6 & 490,571 & 0.2557 $\pm$ 0.0032 &  0.6630 $\pm$ 0.0085\\
    GCN + Transformer & 6 & LapPE &8 & 490,571 & 0.2544 $\pm$ 0.0037 &  0.6593 $\pm$ 0.0060\\
    \midrule
    GCN + GEANet   & 6 & LapPE &2 & 626,219 & 0.2474 $\pm$ 0.0006 &  0.6828 $\pm$ 0.0059\\
    GCN + GEANet   & 6 & LapPE &3 & 539,123 & 0.2461 $\pm$ 0.0006 &  0.6890 $\pm$ 0.0060\\
    GCN + GEANet   & 6 & LapPE &4 & 508,571 & 0.2455 $\pm$ 0.0009 &  0.6892 $\pm$ 0.0042\\
    GCN + GEANet   & 6 & LapPE &6 & 486,683 & 0.2470 $\pm$ 0.0025 &  0.6880 $\pm$ 0.0025\\
    GCN + GEANet   & 6 & LapPE &8 & 463,211 & {\bf0.2445 $\pm$ 0.0013} & {\bf0.6912 $\pm$ 0.0012}\\
    \bottomrule
    \end{tabular}
    }
\end{table*}
\paragraph{Impact of Positional Encoding.} 
Similar to the experiments on attention heads, we study the impact of different positional encodings on attention. Table~\ref{tab:pes} shows the results averaged over 4 different random seeds. We find that GEANet has a lower dependency on positional encoding compared to Transformer with self-attention.
\begin{table*}[ht]
    \vspace{-0.1in}
    \centering 
    \caption{The results of Transformer and GEANet with different positional encodings.}
    \label{tab:pes}
    \fontsize{8.25pt}{8.25pt}\selectfont
    \setlength\tabcolsep{6.25pt} 
    \scalebox{0.9}{
    \begin{tabular}{p{2.5cm}cccccc} 
    \toprule
         {\bf Model} & {\bf \#Layers} & {\bf Positional} & {\bf \#Heads} & {\bf \#Parameters} &{\bf Peptides-Struct } & {\bf Peptides-Func } \\
         &        &     {\bf    Encoding }&        & &MAE $\downarrow$ &AP $\uparrow$   \\

         \midrule
    GCN + Transformer & 6 & None &4 & 492,731 & 0.3871 $\pm$ 0.0094 &  0.6404 $\pm$ 0.0095\\
    GCN + Transformer & 6 & RWPE &4 & 490,143 & 0.2858 $\pm$ 0.0044 &  0.6564 $\pm$ 0.0122\\
    GCN + Transformer & 6 & LapPE &4 & 490,571 & 0.2524 $\pm$ 0.0017 &  0.6589 $\pm$ 0.0069\\
    \midrule
    GCN + GEANet   & 6 & None  &4 & 510,731 & 0.2512 $\pm$ 0.0003 &  0.6722 $\pm$ 0.0065\\
    GCN + GEANet   & 6 & RWPE  &4 & 508,143 & 0.2546 $\pm$ 0.0018 &  0.6794 $\pm$ 0.0089\\
    GCN + GEANet   & 6 & LapPE &4 & 508,571 & {\bf0.2445 $\pm$ 0.0013} &  {\bf0.6892 $\pm$ 0.0042}\\
    \bottomrule
    \end{tabular}
    }
\end{table*}
\paragraph{GEAET in Link Prediction Task.}
In the link prediction task, we evaluate common ranking metrics from the knowledge graph link prediction literature~\citep{hits} as shown in Table~\ref{tab:GEAET_pcqm}: Hits@1, Hits@3, Hits@10 and Mean Reciprocal Rank (MRR), where Hits@$k$ indicates whether the actual answer is among the top-$k$ predictions provided by the model.
\begin{table*}[ht]
    \centering
    \caption{Performance of GEAET on the link prediction task of the PCQM-Contact dataset. We select baseline models that also report the Hits@1, Hits@3, Hits@10, and MRR metrics. Our results are averaged over 4 runs with 4 different seeds, while the results of the baseline models are either from~\citep{Lrgb} or the original papers.}
    \label{tab:GEAET_pcqm}
    \fontsize{8.5pt}{8.5pt}\selectfont
    \setlength\tabcolsep{6.25pt} 
    \scalebox{1}{
    \begin{tabular}{p{2.5cm}cccc} 
    \toprule
         {\bf Model} & {Test Hits@1 $\uparrow$} & {Test Hits@3 $\uparrow$} & {Test Hits@10 $\uparrow$} & {Test MRR $\uparrow$} \\
    \midrule
    SAN             & 0.1312 $\pm$ 0.0016 & 0.4030 $\pm$ 0.0008 & 0.8550 $\pm$ 0.0024 &  0.3341 $\pm$ 0.0006 \\
    GatedGCN        & 0.1288 $\pm$ 0.0013 & 0.3808 $\pm$ 0.0006 & 0.8517 $\pm$ 0.0005 &  0.3242 $\pm$ 0.0008 \\
    Transformer     & 0.1221 $\pm$ 0.0011 & 0.3679 $\pm$ 0.0033 & 0.8517 $\pm$ 0.0039 &  0.3174 $\pm$ 0.0020 \\
    GCN             & 0.1321 $\pm$ 0.0007 & 0.3791 $\pm$ 0.0004 & 0.8256 $\pm$ 0.0006 &  0.3234 $\pm$ 0.0006 \\
    GINE            & 0.1337 $\pm$ 0.0013 & 0.3642 $\pm$ 0.0043 & 0.8147 $\pm$ 0.0062 &  0.3180 $\pm$ 0.0027 \\
    Graph Diffuser  & 0.1369 $\pm$ 0.0012 & 0.4053 $\pm$ 0.0011 & 0.8592 $\pm$ 0.0007 &  0.3388 $\pm$ 0.0011 \\
    \midrule
    GEAET (ours)      & \textbf{0.1566 $\pm$ 0.0014} & \textbf{0.4227 $\pm$ 0.0022} & \textbf{0.8626 $\pm$ 0.0032} &  \textbf{0.3518 $\pm$ 0.0011} \\
    \bottomrule
    \end{tabular}
    }
\end{table*}
\paragraph{Improving GNN with GEANet.}
To demonstrate the importance of GEANet, we conduct additional experiments on the Peptides-Struct, Peptides-Func, and PascalVOC-SP datasets. As shown in Table~\ref{tab:appendix_gnn}, compare with positional encoding, GEANet significantly improves the performance of all base message-passing GNNs.
\begin{table*}[ht]
    \centering
    \caption{Improving GNN performance with GEANet. We run the experiments with 4 different seeds and average the results.}
    \label{tab:appendix_gnn}
    \fontsize{8.25pt}{8.25pt}\selectfont
    \setlength\tabcolsep{6.25pt} 
    \scalebox{0.9}{
    \begin{tabular}{p{2.7cm}ccccc} 
    \toprule
         {\bf Model}&  {\bf Positional} & {\bf PascalVOC-SP}    &{\bf Peptides-Struct }         & {\bf Peptides-Func } \\
                &  {\bf    Encoding }   &  F1 score $\uparrow$ & MAE $\downarrow$     &  AP $\uparrow$ \\
    \midrule
    GCN                 & None  &  0.1268 $\pm$ 0.0060          &  0.3496 $\pm$ 0.0013          &  0.5930 $\pm$ 0.0023 \\
    GCN + GEANet       & None  &  0.2250 $\pm$ 0.0103          &  0.2512 $\pm$ 0.0003          &  0.6722 $\pm$ 0.0065 \\
    GCN + GEANet       & LapPE &  {\bf0.2353 $\pm$ 0.0070}           &  {\bf0.2445 $\pm$ 0.0013}     &  {\bf0.6892 $\pm$ 0.0042} \\
    GCN + GEANet       & RWPE  &  0.2325 $\pm$ 0.0165           &  0.2546 $\pm$ 0.0018          &  0.6794 $\pm$ 0.0089 \\
    \midrule
    GINE                & None  &  0.1265 $\pm$ 0.0076          &  0.3547 $\pm$ 0.0045          &  0.5498 $\pm$ 0.0079 \\
    GINE + GEANet      & None  &  0.2742 $\pm$ 0.0032          &  0.2544 $\pm$ 0.0012          &  0.6509 $\pm$ 0.0021 \\
    GINE + GEANet      & LapPE &  0.2746 $\pm$ 0.0071           &  {\bf0.2480 $\pm$ 0.0023}     &  {\bf0.6654 $\pm$ 0.0055} \\
    GINE + GEANet      & RWPE  &  {\bf0.2762 $\pm$ 0.0022}           &  0.2546 $\pm$ 0.0011          &  0.6618 $\pm$ 0.0059 \\
    \midrule
    GatedGCN            & None  &  0.2873 $\pm$ 0.0219          &  0.3420 $\pm$ 0.0013          &  0.5864 $\pm$ 0.0077 \\
    GatedGCN + GEANet  & None  &  0.3933 $\pm$ 0.0027          &  0.2547 $\pm$ 0.0009          &  0.6485 $\pm$ 0.0035\\
    GatedGCN + GEANet  & LapPE &  {\bf0.3944 $\pm$ 0.0044}           &  {\bf0.2468 $\pm$ 0.0014}     &  0.6715 $\pm$ 0.0034 \\
    GatedGCN + GEANet  & RWPE  &  0.3899 $\pm$ 0.0017           &  0.2577 $\pm$ 0.0006          &  {\bf0.6734 $\pm$ 0.0028} \\
    \bottomrule
    \end{tabular}
    }
\end{table*}
% \section{Complexity Analysis}  \label{sec:complexity}
\section{Complexity Analysis}  \label{sec:complexity}
We first analyze the complexity of GEANet.
As the model dimensions $d$ and the size of external units $S$ are hyper-parameters, GEANet scales linearly with the number of nodes and edges, resulting in a complexity of $O(|\Vcal|+|\Ecal|)$. 
For GEAET, the complexity is primarily determined by GEANet, Transformer and message-passing GNN. 
The GEANet, as described above, has linear complexity.
The message-passing GNN has a complexity of $O(|\Ecal|)$. In typical cases, the Transformer uses the self-attention mechanism with a complexity of $O(|\Vcal|^2)$, resulting in complexity of $O(|\Vcal|^2)$ .
In practice, we observe that on certain datasets such as Peptides-Struct and Peptides-Func, not using Transformer yields better results, achieving linear complexity in such cases. 
Additionally, we can use linear Transformers to reduce the complexity of GEAET to linearity.
\section{Model Interpretation} \label{sec:appendix_interpretation}
\begin{figure}[htbp]
    \includegraphics[width=.2\textwidth]{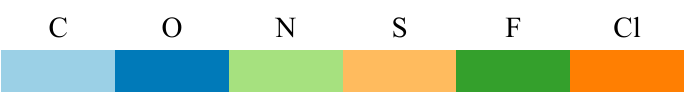} \\
    \begin{center}
    \includegraphics[width=.47\textwidth]{Figures/attn/graph821_title.pdf}      % 2.122,2.054,1.675
    \hspace{0.5cm}
    \includegraphics[width=.47\textwidth]{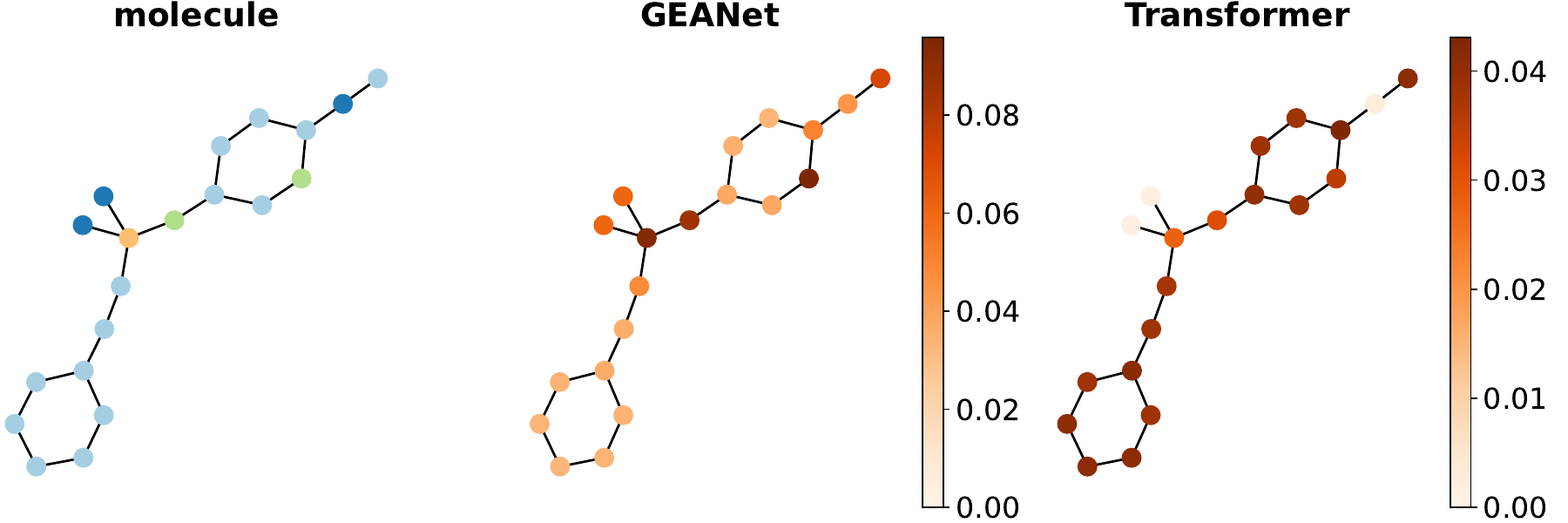} \\    % 1.202,1.047,1.514
    \includegraphics[width=.47\textwidth]{Figures/attn/graph277.pdf}            % 0.298,0.476,0.538
    \hspace{0.5cm}
    \includegraphics[width=.47\textwidth]{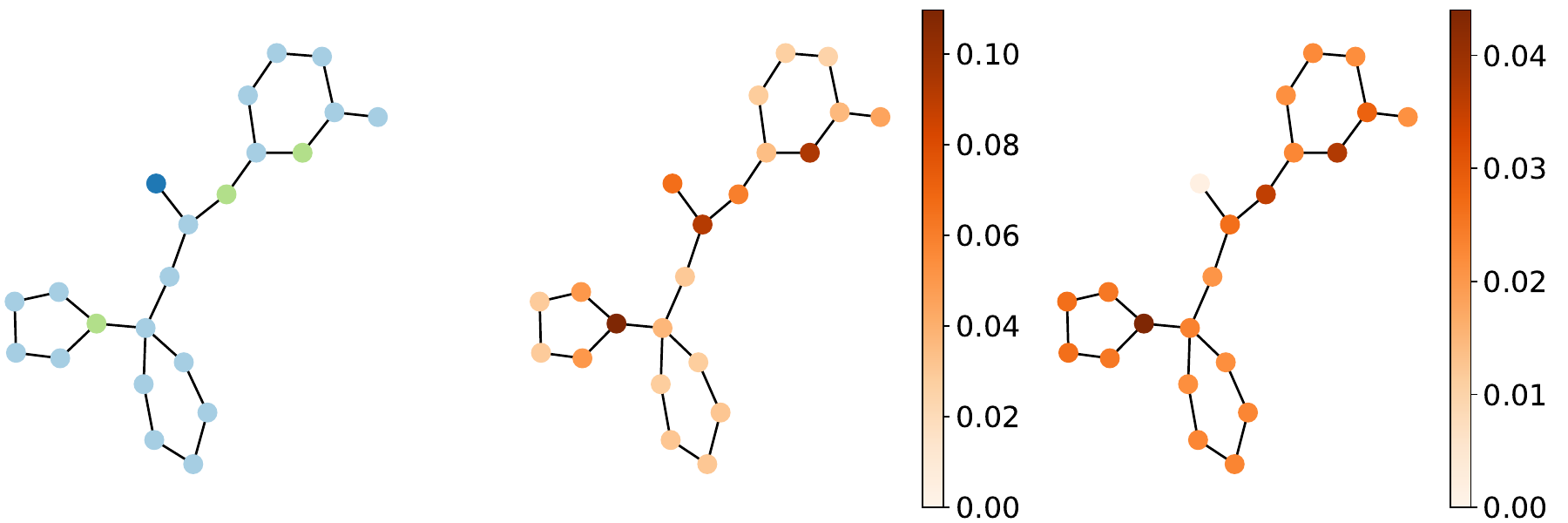} \\         % 1.493,1.732,2.526
    \includegraphics[width=.47\textwidth]{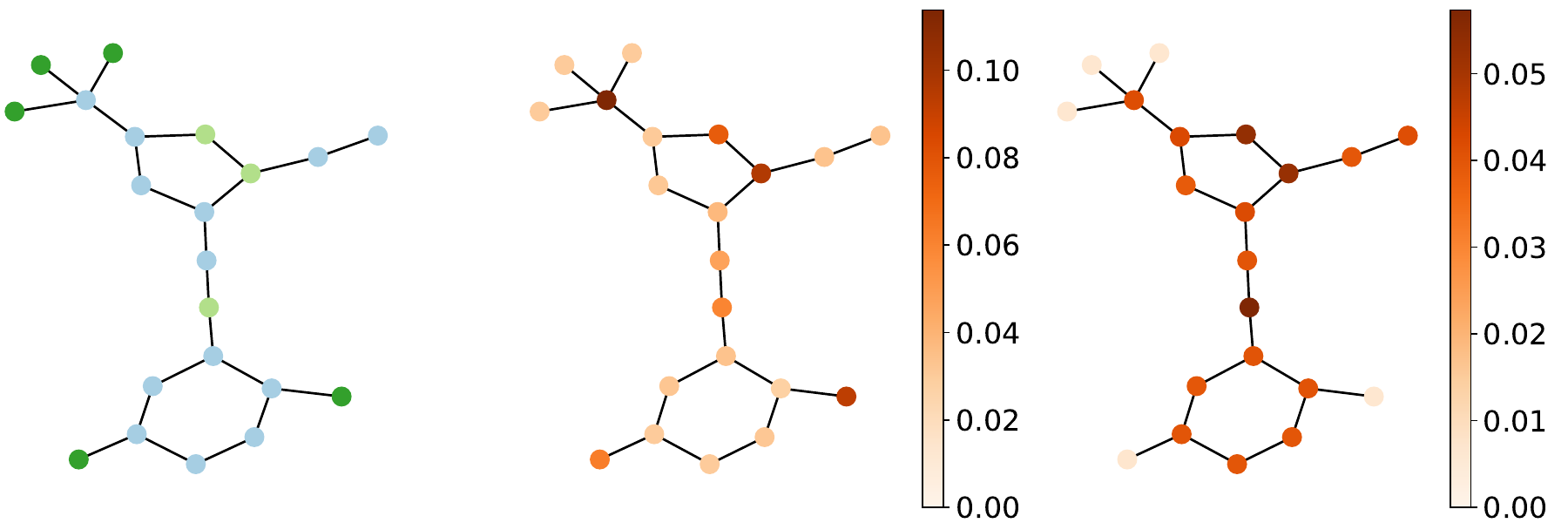}            % 1.917,1.976,1.796
    \hspace{0.5cm}
    \includegraphics[width=.47\textwidth]{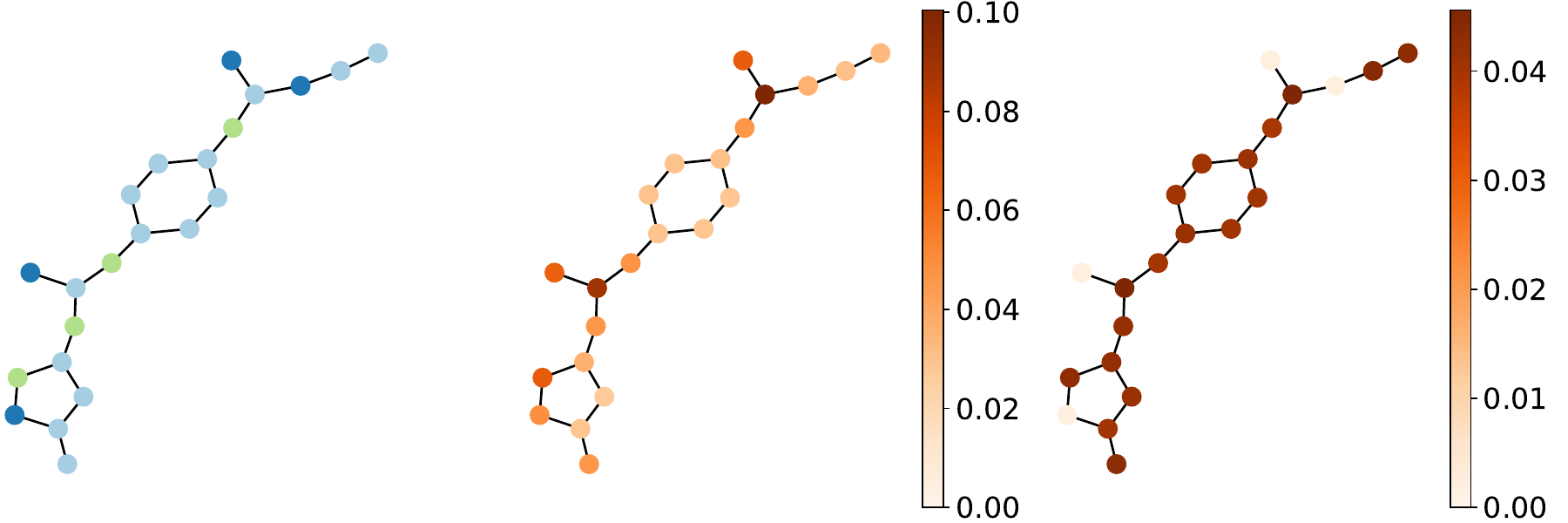} \\         % 1.991,1.839,0.681
    \includegraphics[width=.47\textwidth]{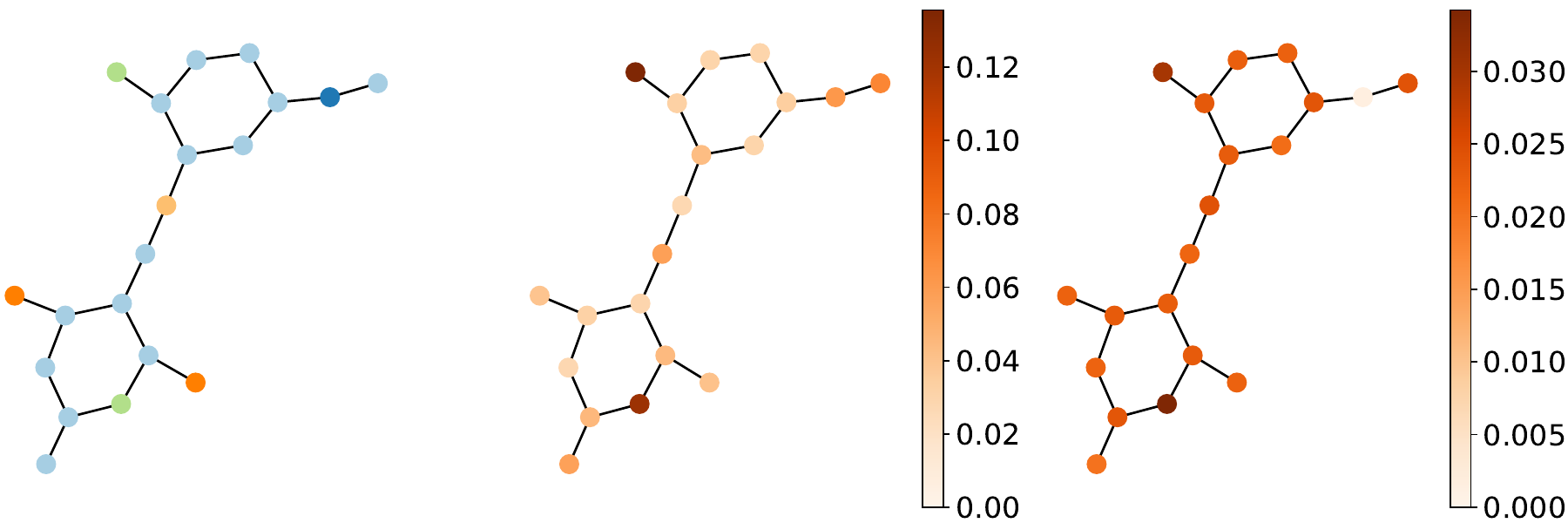}            % 2.322,2.331,2.941
    \hspace{0.5cm}
    \includegraphics[width=.47\textwidth]{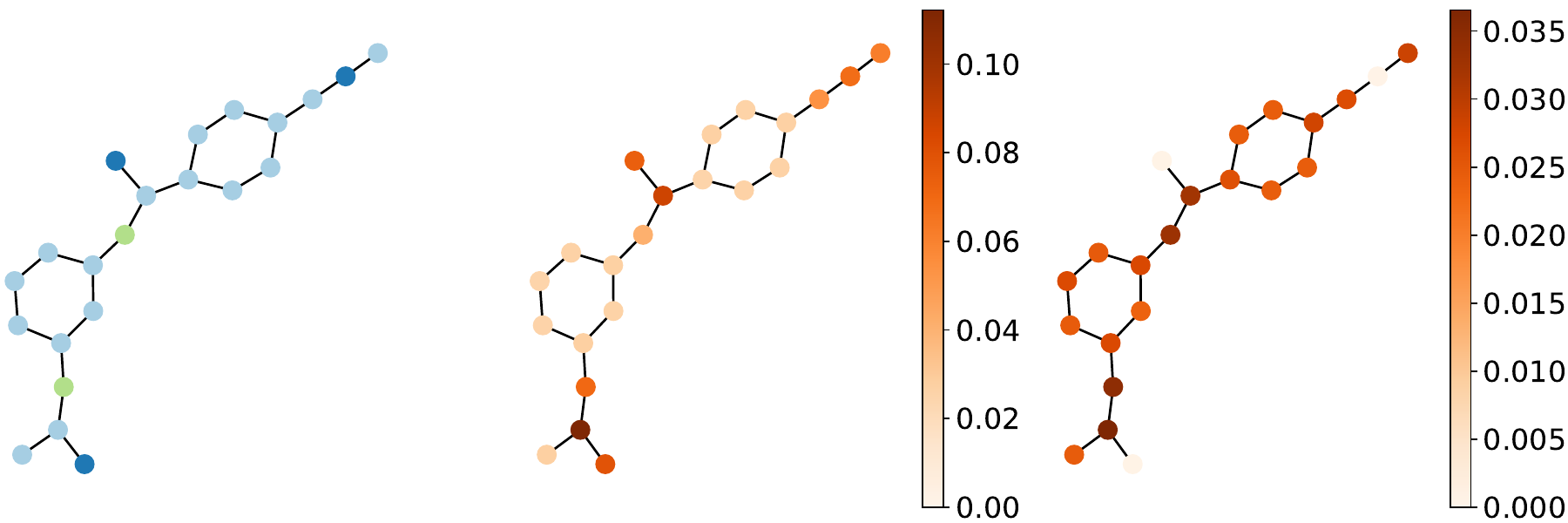} \\         % 2.311,2.299,2.217
    \caption{Attention visualization of GEANet and Transformer on ZINC molecular graphs. The center column shows the attention weights of GEANet and the right column shows the attention weights learned by the classic Transformer.}
    \label{fig:appendix_visual}
    \end{center}
\end{figure}

In addition to the examples presented in the main paper, we provide additional visualization results in Figure~\ref{fig:appendix_visual}. 
GEANet and Transformer use the same positional encoding, and other hyperparameter settings are generally consistent. 
The first column shows the original molecules from ZINC, the middle and right columns show the visualization results of GEANet and Transformer, respectively. 
We observe that GEANet focuses more on the important nodes or connected nodes of specific structures, which improves the ability to distinguish different graphs or motifs. 
This indicates that GEANet excels in handling structural information and concentrates on discriminative nodes.

% You can have as much text here as you want. The main body must be at most $8$ pages long.
% For the final version, one more page can be added.
% If you want, you can use an appendix like this one.  

% The $\mathtt{\backslash onecolumn}$ command above can be kept in place if you prefer a one-column appendix, or can be removed if you prefer a two-column appendix.  Apart from this possible change, the style (font size, spacing, margins, page numbering, etc.) should be kept the same as the main body.
%%%%%%%%%%%%%%%%%%%%%%%%%%%%%%%%%%%%%%%%%%%%%%%%%%%%%%%%%%%%%%%%%%%%%%%%%%%%%%%
%%%%%%%%%%%%%%%%%%%%%%%%%%%%%%%%%%%%%%%%%%%%%%%%%%%%%%%%%%%%%%%%%%%%%%%%%%%%%%%

\end{document}